\documentclass[12pt, letterpaper]{amsart}
\usepackage[left=1in,right=1in,bottom=0.5in,top=0.8in]{geometry}
\usepackage{graphicx,subcaption,lipsum}
\usepackage{amsfonts}
\usepackage{amsfonts,amsbsy,amssymb,amsmath,amsthm, bm} 
\usepackage[font=small,labelfont=bf]{caption}
\usepackage{epstopdf}
\usepackage{xcolor}

\usepackage{float}
\usepackage{pgfplots}
\usepackage{listings}
\usepackage{longtable}
\usepackage{mathrsfs}
\usepackage{mathtools}
\usepackage{algorithm,algorithmic}
\usepackage{textcomp}
\usepackage{stfloats}
\usepackage{url}
\usepackage{verbatim}
\usepackage{xcolor}

\usepackage{epsfig}
\usepackage{dsfont}
\usepackage{natbib}

\newtheorem{thm}{Theorem}[section]

\newtheorem{assumption}[thm]{Assumption}

\theoremstyle{definition}

\newtheorem{rem}[thm]{Remark}

\definecolor{energy}{RGB}{114,0,172}
\definecolor{freq}{RGB}{45,177,93}
\definecolor{spin}{RGB}{251,0,29}
\definecolor{signal}{RGB}{203,23,206}
\definecolor{circle}{RGB}{217,86,16}
\definecolor{average}{RGB}{203,23,206}

\colorlet{shadecolor}{gray!20}
\pgfplotsset{compat=1.9}

\usepgflibrary{fpu}

\newcommand{\argmin}{\mathop{\mathrm{arg\,min}}}

\def\1{1\kern-.20em {\rm l}}
\usepackage{enumitem}
\usepackage{lscape}
\usepackage{booktabs}

\numberwithin{equation}{section}

\author[Mohamed Chaouch]{Mohamed Chaouch}
\address{(M. Chaouch) Statistics Program, Department of Mathematics and Statistics, College of Arts and Sciences, Qatar University, Doha, Qatar}
\email[Corresponding author]{mchaouch@qu.edu.qa}
\urladdr{}
\author[Omama M. Al-Hamed]{Omama M. Al-Hamed}
\address{(O. M. Al Hamed) Statistics Program, Department of Mathematics and Statistics, College of Arts and Sciences, Qatar University, Doha, Qatar}
\email{omama5854@gmail.com}
\urladdr{}
\title[Online Nonparametric Supervised Learning for Massive Data]{Online Nonparametric Supervised Learning for Massive Data}
\begin{document}
\begin{abstract}
Despite their benefits in terms of simplicity, low computational cost and data requirement, parametric machine learning algorithms, such as linear discriminant analysis, quadratic discriminant analysis or logistic regression, suffer from serious drawbacks including linearity, poor fit of features to the usually imposed normal distribution and high dimensionality. Batch kernel-based nonparametric classifier, which overcomes the linearity and normality of features constraints, represent an interesting alternative for supervised classification problem. However, it suffers from the ``curse of dimension". The problem can be alleviated by the explosive sample size in the era of big data, while large-scale data size presents some challenges in the storage of data and the calculation of the classifier. These challenges make the classical batch nonparametric classifier no longer applicable. This motivates us to develop a fast algorithm adapted to the real-time calculation of the nonparametric classifier in massive as well as streaming data frameworks. This online classifier includes two steps. First, we consider an online principle components analysis to reduce the dimension of the features with a very low computation cost. Then, a stochastic approximation algorithm is deployed to obtain a real-time calculation of the nonparametric classifier. The proposed methods are evaluated and compared to some commonly used machine learning algorithms for real-time fetal well-being monitoring. The study revealed that, in terms of accuracy, the offline (or Batch), as well as, the online classifiers are good competitors to the random forest algorithm. Moreover, we show that the online classifier gives the best trade-off accuracy/computation cost compared to the offline classifier.


\end{abstract}
\maketitle

\section{Introduction}\label{intro}

\subsection{Supervised learning in the era of big data}\label{PB}
Supervised learning algorithms represent a powerful tool for practitioners to help in taking decisions. For instance in banking sector it is important that credit card companies recognize fraudulent credit card transactions so that customers are not charged for items that they did not purchase (see \cite{Al}). A real-time detection of such fraud allows the bank to take the appropriate measures to block the credit card. In healthcare domain, \cite {Ba} considered real-time monitoring of fetal health during gestation period is important to reduce the mortality risk of the fetus. Given the cardiotocographic data, supervised learning techniques may predict and label the health state of the fetus as normal, needs guarantee, or pathology. Therefore, they represent a preventive tool to the medical expert in case of high risk situations. In food quality control field, usually companies producing meat, for instance, are interested in assessing the quality of their meat to stay competitor in the market. Based on some biological, texture or color features of the meat, one can use supervised learning techniques to predict the quality of the meat without going through a costly and time consuming chemical and biological analysis procedures. The reader is referred to \cite{Zhu}, and the references therein, for a survey on deep learning and machine vision techniques used for food processing. For a further discussion on the use of machine learning algorithms to solve real-world decision making problems one refers to \cite{EH} and \cite{Sarker}.

Nowadays with the progress in the technology of electronic devises, we have an easy access to a vast amounts of structured, semi-structured and unstructured information. Supervised learning algorithms play a crucial role in extracting valuable insights, detecting patterns, and making accurate predictions and decisions. The five-Vs defining the big data setting, namely Volume, Variety, Velocity, Veracity and Value, make the application of most of supervised learning approaches to massive data very challenging. Indeed, it is difficult to deal with using traditional learning methods since the established process of learning from conventional datasets was not designed to and will not work well with high volumes of data. For instance, \cite{Chen} reported that most traditional machine learning algorithms are designed for data that would be completely loaded into memory which does not hold any more in the context of big data. Another assumption on which are based most of the classical machine learning algorithms is that the entire dataset is available for processing at the time of training. However, in real life, data are frequently received in streaming fashion and all the data are not necessarily stored at once. In such case, real-time cleaning, processing and understanding of such streaming data content becomes of great interest for a fast and accurate decision making. \cite{Heureux} and \cite{Qiu} have discussed how Big Data framework did break these assumptions, rendering traditional algorithms unstable or greatly impeding their performance.

\subsection{Related work} 
In this subsection, we aim to remind the reader about the statistical formulation of the supervised learning problem. Then, we briefly revisit the classical supervised learning methods and discuss their major drawback in order to motivate the introduction of nonparametric supervised techniques in the next section. We also discuss methods introduced in the literature to extend supervised learning algorithms to the big data setting.

In a supervised learning problem, we suppose that we have a set of observed features, say $\mathbf{X}\in \mathbb{R}^d$ ($d\geq 1$), that may explain and allow to predict a certain categorical variable, say $Y$, that takes values in a set of $G$ known classes denoted $\mathcal{G}:=\{ 1, \dots, G\}.$ Suppose that the available data $(\mathbf{x}_1, y_1), \dots, (\mathbf{x}_n, y_n)$ are realizations of the pair of random variables $(\mathbf{X},Y)\in \mathbb{R}^d\times\mathcal{G}$ with unknown probability distribution. The main task in supervised learning consists in finding a classification rule (classifier) denoted $m: \mathbb{R}^d \rightarrow \mathcal{G}$ such that for each input vector of features $\mathbf{x}\in \mathbb{R}^d$, it returns a prediction of a class $m(\mathbf{x})\in \mathcal{G}.$

Given a classifier $m$, its performance is measured via the misspecification rate (classification error) defined, for any fixed $\mathbf{x}$, as 
$\mathcal{L}(m) = \mathbb{E}\left( \1_{\{m(\mathbf{X})\neq Y \}}| \mathbf{X}=\mathbf{x}\right) = \mathbb{P}\left(m(\mathbf{X})\neq Y | \mathbf{X}=\mathbf{x} \right).$
Therefore, the problem in supervised learning turns to build a classifier that minimizes $\mathcal{L}(\cdot).$ Note that the Bayes' rule allows to say that a classifier $m^\star: \mathbb{R}^d \rightarrow \mathcal{G}$ defined, for a fixed $\mathbf{x}$, as 
$
m^\star(\mathbf{x}) = \arg\max_{g\in \mathcal{G}} \mathbb{P}\left( Y=g | \mathbf{X}=\mathbf{x}\right)
$
is optimal in the sense that $\mathcal{L}(m^\star) \leq \mathcal{L}(m)$, for any classifier $m.$ This means that, given a new item with vector of features $\mathbf{x}$, the classifier $m^\star$ will assign to it the class $g$ corresponding to the highest conditional probability $\mathbb{P}\left( Y=g | \mathbf{X}=\mathbf{x}\right).$ Consequently, the major task now consists in estimating this conditional probability to have an optimal classifier that can be easily used in practice.

Several supervised learning methods were introduced in the literature. For a general overview of several supervised and unsupervised learning techniques, the reader is referred to e.g. \cite{Hastie}. Let us now revisit some known supervised learning algorithms that will be used as benchmark to the method we introduce in the following section.
\subsubsection{Review of Linear Discrimination Analysis (LDA)}
Let $f_g(\mathbf{x})$, for $g \in \{1, \dots, G \}$, be the multivariate conditional density of $\mathbf{X}$ given $Y=g.$ We denote by $\pi_g:= \mathbb{P}(Y=g)$ the prior probability that $Y$ is in the class $g.$ Then, using the Bayes' theorem one obtains the following posterior probability
\begin{eqnarray}\label{LDAprob}
\mathbb{P}(Y=g | \mathbf{X}=\mathbf{x}) = \dfrac{f_g(\mathbf{x}) \pi_g}{\sum_{\ell=1}^G \pi_\ell f_\ell(\mathbf{x})}.
\end{eqnarray}
Note that all probability distributions in \eqref{LDAprob} are usually unknown. Therefore, it is easy to see that a good estimation of the conditional density of $\mathbf{X}|Y=g$ leads to a good estimation of $\mathbb{P}(Y=g | \mathbf{X}=\mathbf{x}).$ The LDA relies on the following assumption:
\begin{assumption}\label{as1}
Conditional on class membership $g\in \mathcal{G}$, the samples ${\bf x}_{g,i} \in \mathbb{R}^d$, for $i=1, \dots, n$, are independent and identically distributed from $\mathcal{N}_d\left(\boldsymbol{\mu}_g, \boldsymbol{\Sigma} \right),$ where $\mathcal{N}_d$ denotes the multivariate Gaussian distribution with vector of means $\boldsymbol{\mu}_g$ and covariance matrix $\boldsymbol{\Sigma}.$
\end{assumption}
 
Under Assumption \ref{as1}, the discriminant frontier function of the class $g$, say $\delta_g(\mathbf{x}),$ between classes could be expressed as a linear function of $\mathbf{x}$ according to the following expression
$
\delta_g(\mathbf{x}) = \mathbf{x}^\top \boldsymbol{\Sigma}^{-1}\boldsymbol{\mu}_g - \frac{1}{2} \boldsymbol{\mu}^\top \boldsymbol{\Sigma}^{-1} \boldsymbol{\mu}_g +\log(\pi_g), 
$
where $\mathbf{x}^\top$ denotes the transpose of the vector $\mathbf{x}$. Therefore, maximizing, with respect to $g\in \mathcal{G},$ the posterior probability $\mathbb{P}(Y=g | \mathbf{X}=\mathbf{x})$ comes to maximize the discriminant frontier function $\delta_g(\mathbf{x}).$ However, since $\boldsymbol{\mu}_g$ and $\boldsymbol{\Sigma}$ are unknown, then they have to be estimated using the available training data. 
\begin{rem}
    Note that the LDA method assumes the normality of the vector of random features $\mathbf{X}$ conditional to $Y$ and that the covariance matrix is constant for all classes $g\in \{1, \dots, G\}.$ This is a very restrictive condition which can easily be violated whenever data are received in streaming fashion since the probability distribution generating features may change shape and/or parameters over time. Moreover, to learn the discriminant frontier function $\delta_g({\bf x})$ a storage of all the data set on the same computer is required to achieve the desired estimation accuracy. In massive data context it might be infeasible to keep the large-scale data set in memory or even store all the data on a single computer when the data size is too large. Recently, \cite{Chu} introduced an incremental linear discriminant analysis that is more adapted the data streaming context. They proposed a new batch LDA algorithm called LDA/QR. LDA/QR is a simple and fast LDA algorithm, which is obtained by computing the economic QR factorization of the data matrix followed by solving a lower triangular linear system. Then, based on LDA/QR, they developed a new incremental LDA algorithm called ILDA/QR which can easily handle the update from one new sample or a chunk of new samples, it has efficient computational complexity and space complexity and it is very fast and always achieves competitive classification accuracy. Moreover, note that the estimation of the covariance matrix $\boldsymbol{\Sigma}$ has a complexity $O(n d^2)$, which makes LDA computationally expensive on large-sample data and high dimensional features. \cite{lap} proposed a compression approach for reducing the number of training samples for LDA and quadratic discriminant analysis. 
\end{rem}
\subsubsection{Quadratic Discriminant Analysis (QDA)}
In contrast to the LDA, the QDA came to consider the non-homogeneity of the covariance matrix between classes while still assuming the normality of features $\mathbf{x}$ under each class $g\in \mathcal{G}$. The QDA is based on the following assumption:
\begin{assumption}\label{as2}
    Conditional on class membership $g\in \mathcal{G}$, the samples ${\bf x}_{g,i} \in \mathbb{R}^d$, for $i=1, \dots, n$, are independent and identically distributed from $\mathcal{N}_d\left(\boldsymbol{\mu}_g, \boldsymbol{\Sigma}_g \right),$ where $\boldsymbol{\Sigma}_g$ is the covariance matrix under the class $g.$
\end{assumption}
Under Assumption \ref{as2}, the discriminant frontier function of the class $g$ can be expressed as a quadratic function of $\mathbf{x}$. That is, for $g \in\mathcal{G},$
$
\delta_g(x)=-\frac{1}{2} \log \left|\boldsymbol{\Sigma}_g\right|-\frac{1}{2}\left(\mathbf{x}-\boldsymbol{\mu}_g\right)^t \boldsymbol{\Sigma}_g^{-1}\left(\mathbf{x}-\boldsymbol{\mu}_g\right)+\log \pi_g,
$
where $\left|\boldsymbol{\Sigma}_g\right|$ denotes the determinant of the matrix $\boldsymbol{\Sigma}_g.$ Then, as in the LDA, for a fixed vector of features $\mathbf{x},$ the optimal class will be
$
Y^{\text{o}} := \arg\max_{g\in \mathcal{G}} \mathbb{P}(Y=g | \mathbf{X}=\mathbf{x}) = \arg\max_{g\in \mathcal{G}} \delta_g(\mathbf{x}).
$
This class is easily predicted as soon as we have a consistent estimation of the posterior probability, or equivalently, an estimation of $\delta_g(\mathbf{x})$ through the estimation of the parameters of the normal distribution $\boldsymbol{\mu}_g$ and $\boldsymbol{\Sigma}_g$ within each class $g\in \mathcal{G}.$ 

\begin{rem}
As for the LDA, the form of the discriminant frontier function is essentially based on the normality assumption of features. Moreover, assuming that such probability distribution remains stationary and does not change over time is not realistic in the context of streaming data for instance. From a computational view point, the estimation of discriminant frontier function requires the storage of all the training sample on the same computer. In the context of massive data, it is very challenging to store a large-scale data set on one machine which makes the estimation of $\delta_g(\mathbf{x})$ infeasible.  
\end{rem}

\subsubsection{$k$-Nearest Neighbors classifier ($k$NN)}
In contrast to LDA and QDA, the $k$-Nearest Neighbors method is a well-known nonparametric supervised learning method that relaxes the normality assumption of the vector of features $\mathbf{X}$ under each class $g\in \mathcal{G}.$

Let $(\mathbf{X}_1, Y_1), \dots, (\mathbf{X}_n, Y_n)$ be a training sample and consider $k\leq n$. Then, the $k$NN method consists in assigning, to the individual with vector of features $\mathbf{x}$, the most likely class of its closest neighbors. The closeness here should be understood according to a certain metric measuring the similarity between $\mathbf{x}$ and the features in the training sample $\mathbf{X}_1, \dots, \mathbf{X}_n.$ More formally, the predicted class will be defined, for a fixed $k$, as 
$
\widehat{Y} = \arg\max_{g\in\mathcal{G}}\sum_{i\in \mathcal{V(\mathbf{x})}} \1_{\{Y_i = g\}},
$
where $\mathcal{V}(\mathbf{x}) = \{i: d(\mathbf{X}_i,\mathbf{x}) \leq \tau\}$ denotes the neighborhood of $\mathbf{x}$ such that $d(\cdot, \cdot)$ is a metric that measures the similarity between $\mathbf{X}_i$ and $\mathbf{x}$ and $\tau$ is a threshold below which we consider the observation $\mathbf{X}_i$ similar enough to $\mathbf{x}.$ It is clear that the $k$NN classifier depends on two tuning parameters which are the number of neighbors $k$ and the metric $d(\cdot, \cdot).$ In practice several distance metrics can be used including the Euclidean distance, Manhattan distance, Minkowski distance and Hamming distance. The choice of the number of neighbors $k$ is very crucial in practice since a high value of $k$ will generate a small variance but high bias. However, a small value of $k$ induces an overfitting, that is a high variance and a small bias. Cross-validation technique is in general used in practice for an optimal choice of $k$ since it secures an optimal trade-off between bias and variance of the $k$NN classifier. 

\begin{rem}
    In contrast to the LDA and QDA, the advantage of the $k$NN algorithm is that it does not rely on any assumed probability distribution for the features. It is a metric-based approach where the class assigned to any new observation, with known feature ${\bf x}$, is based on major votes among all observations in its neighborhood. However, the $k$NN algorithm requires the storage of all data in the training sample in order to calculate all distances between observations. Moreover, the optimal number of neighbors $k$ is usually obtained by cross-validation technique which requires the storage of all training sample on one computer. In the context of big data, in addition to the challenge of saving large-scale data at once on the same machine, the use of cross-validation technique to find the optimal $k$ makes the $k$NN approach less efficient from a computational view point. The computation complexity for the $k$NN algorithm is $O(nkd)$ which clearly increases with the training sample size as well as the dimension of features.   
\end{rem}

There is a variety of supervised learning techniques in the literature. However, we will limit our review here to these three methods as benchmarks to be compared to the new nonparametric classifiers introduced in this paper. Other advanced algorithm such as Random Forest (RF) will also be considered for comparison purpose.

\subsection{Contribution of the paper}

The contribution of this paper is two folds: 
\begin{enumerate}
\item First, we introduce an offline nonparametric classifier based on kernel-type estimator of the posterior probability $\mathbb{P}\left(Y=g | \mathbf{X}=\textbf{x}\right).$ Locally weighted nonparametric estimators represent an interesting alternative to parametric learning models such as the LDA and QDA. Indeed, they do not require any assumption on the probability distribution of the pair $(\mathbf{X},Y)$ and the classifier is directly learned from the data. The offline classifier suffers from the well-known ``curse of dimension". In other words higher is the dimension of features, less accurate will be the kernel-type classifier (for further discussion see subsection \ref{cd}). For this, a Batch Principle Component Analysis (PCA) is used to first reduce the dimension of the space of features. Note that in the context of big data, the sample size is extremely large which helps in reducing the effect of the curse of dimension problem on the accuracy of the classifier. However, it may be infeasible to keep the large-scale data set in memory or even store all the data on a single computer when the data size is too large. This makes the application of the offline classifier very challenging to apply in case of massive data context.
\item The second contribution of this paper consists in using stochastic approximation algorithm to define an online nonparametric classifier that estimates the posterior probability $\mathbb{P}\left(Y=g | \mathbf{X}=\textbf{x}\right)$ in real-time without necessarily storing all the large-scale data set on a single computer. Moreover, to handle the high dimension of the features, we consider an online PCA where the estimate of principle components is updated every time a new observation is received in the data base. The combination of the online PCA as dimension reduction tool and the recursive estimation of the posterior probability allow to obtain an efficient classifier adapted to the case of big data as well as data streaming.
\end{enumerate}

\subsection{Outline of the paper}
The rest of this paper is organized as follows. In section \ref{method} we describe the proposed offline and online classifiers a long with the Batch and Online PCA for the dimension reduction step. In section \ref{app} we present an application of the proposed classifiers to real-time monitoring of fetal well-being during pregnancy. The performance of the proposed classifiers is compared to some commonly used supervised classification algorithms such as Random Forest and $k$NN.

\section{The proposed methodology}\label{method}

\subsection{Offline nonparametric supervised learning}
Let $Y$ be a categorical variables that takes values in one of $G$ possible classes, and $\mathbf{X}=(X_1, \dots, X_d)^\top$ be a vector of $d$ predictors that may explain the class to which belong $Y.$ Suppose that a training sample of $n$ observations of the pair $(\mathbf{X},Y)$ is available, say $(\mathbf{X}_1, Y_1), \dots, (\mathbf{X}_n,Y_n).$ Our purpose is to predict the class of an out-of-sample (new) observation given the values of its vector of predictors $\mathbf{X}$. More, explicitly, given $\mathbf{X} = \mathbf{x}_{\text{new}}$, we calculate $\mathbb{P}\left(Y_{\text{new}} = g | \mathbf{X}=\mathbf{x}_{\text{new}}\right)$, for $g = 1, \dots, G.$ The class assigned to $Y_{\text{new}}$ will be the one corresponding to the highest probability. Therefore, our purpose now consists in estimating nonparametrically $\mathbb{P}\left(Y_{\text{new}} = g | \mathbf{X} = \mathbf{x}_{\text{new}}\right).$

\subsubsection{Offline nonparametric classifier estimation}

\noindent Note that, $\1_{\{Y_{\text{new}}=g\}}$ might be seen as a Bernoulli random variable with $\mathbb{P}(Y_{\text{new}}=g)$ being the probability of success. Therefore, for any $g\in \mathcal{G},$ one has
\vskip-5mm
\begin{eqnarray}
    \label{proba}
    \mathbb{P}_g(\textbf{x}_{\text{new}}) := \mathbb{P}(Y_{\text{new}}=g|\mathbf{X} = \mathbf{x}_{\text{new}})=\mathbb{E}(\1_{\{Y_{\text{new}}=g\}}|\mathbf{X} = \mathbf{x}_{\text{new}}).
\end{eqnarray}
The estimation of a conditional expectation, when the response variable $Y$ is a continuous random variable, is a well-known problem in statistics. Several estimators were proposed in the literature (see \cite{FG} for more details). In this paper, our response variable is categorical. Therefore, we extend the Nadaraya-Watson (see \cite{N} and \cite{W}) estimator of the conditional expectation in \eqref{proba} to the case where $Y$ is categorical to introduce an offline supervised classifier. Note that the computation of this classifier requires the storage of all massive data on the same computer to be able to perform the training as well as  prediction part. For this reason it is called offline (or Batch) classifier.
\vskip-5mm
\begin{eqnarray}
  \label{p_est}
  \widehat{\mathbb{P}}_{g,n}(\mathbf{x}_{\text{new}})=\dfrac{\sum_{i=1}^n\1_{\{Y_i=g\}}K\left(h_g^{-1}\|\mathbf{X}_i-\mathbf{x}_{\text{new}}\|\right)}{\sum_{i=1}^n K\left(h_g^{-1}\|\mathbf{X}_i-\mathbf{x}_{\text{new}}\|\right)}, \quad \forall g\in \mathcal{G},
\end{eqnarray}
where $K: \mathbb{R}^d \rightarrow \mathbb{R}$ is the kernel and $h_g:= (h_{n,g})_{n\in \mathbb{N}}$ is the bandwidth, which is a sequence of positive numbers tending to zero as $n$ goes to infinity. 

\subsubsection{Curse of dimension}\label{cd}

The definition of offline classifier defined in \eqref{p_est} suffers from the well-known ``curse of dimension'' problem. Indeed, \cite{Bosq} (page 70)  have shown that, under some regularity condition on the regression function, and if we suppose that $h_n = c_n n^{-1/(d+4)}$, where $c_n$ is a sequence that converges to $c$ as $n$ goes to infinity, then asymptotically one has $\mathbb{E}[(\widehat{\mathbb{P}}_{g,n}(\mathbf{x}_{\text{new}}) - \mathbb{P}_g(\textbf{x}_{\text{new}}))^2] = c\, n^{-4/(d+4)}.$ This means that the mean square convergence rate of $\widehat{\mathbb{P}}_{g,n}(\mathbf{x}_{\text{new}})$ towards $\mathbb{P}_g(\textbf{x}_{\text{new}})$ depends on the size of the learning sample $n$ as well as the dimension $d$ of the vector of features $\mathbf{X}.$ To clearly see the curse of dimension effect of the quality of estimation, let us consider $n$ fixed and vary $d$. One can observe that, if $d=1$ the convergence rate is proportional to $n^{-4/5}$ and when $d=2$ it becomes proportional to $n^{-4/6}.$ In other words, higher is the dimension $d$, slower will be the convergence rate. 

\begin{rem}
In order to have a reasonable convergence rate of the offline classifier, we either need to reduce the dimension $d$ of the vector of predictors $\mathbf{X}$ or significantly increase the size $n$ of the training sample. Unfortunately, the second option is not always possible to achieve since in many applications (especially in experimental fields such as biology, chemistry, health science or medicine) increasing the sample size might be very costly. In such case, considering a dimension reduction technique to reduce the dimension of features becomes a must in order to improve the accuracy of the offline classifier. However, in the case of big data the sample size is extremely large which makes it possible to achieve the desired accuracy of the offline classifier even when the dimension of features is large. In such case the dimension reduction step is not necessary. 
\end{rem}

\subsubsection{Dimension reduction via Batch PCA}
In this subsection we briefly remind the reader about Batch PCA as a dimension reduction tool for a better understanding of the online PCA which will be discussed in Section \ref{onlinesup}.

Given data points in the form $\mathbb{X} = \left(\mathbf{X}_1, \dots, \mathbf{X}_n \right)\in \mathcal{M}_{d,n}$, where each vector $\mathbf{X}_k$ is $d$-dimensional and possibly large. The goal of batch PCA is to describe/visualize this data in a low-dimensional subspace of $\mathbb{R}^d$. Intuitively, we want to: (1) find a $q$-dimensional affine subspace on which the projected points are the best approximations of the original data, (2) find the projection preserving as much as possible the variance of the original data.
The quality of the representation is measured by the squared distance between the (centered) vectors and their projections in the subspace. Denoting the sample mean by $\boldsymbol{\mu}_n=\frac{1}{n} \sum_{i=1}^n \mathbf{X}_i$, the goal of batch PCA is thus to find a projection matrix $\mathbf{P}_q$ of rank $q \leq d$ that minimizes the loss function
$
\mathcal{R}_n\left(\mathbf{P}_q\right)=\frac{1}{n} \sum_{i=1}^n\left\|\left(\mathbf{X}_i-\boldsymbol{\mu}_n\right)-\mathbf{P}_q\left(\mathbf{X}_i-\boldsymbol{\mu}_n\right)\right\|^2
$
Consider the sample covariance matrix
$
\boldsymbol{\Sigma}_n=\frac{1}{n} \sum_{i=1}^n\left(\mathbf{X}_i-\boldsymbol{\mu}_n\right)\left(\mathbf{X}_i-\boldsymbol{\mu}_n\right)^\top.
$
Let $\mathbf{u}_{1, n}, \ldots, \mathbf{u}_{d, n}$ be orthonormal eigenvectors of $\boldsymbol{\Sigma}_n$ with associated eigenvalues $\lambda_{1, n} \geq \ldots \geq$ $\lambda_{d, n} \geq 0$. The minimum of $\mathcal{R}_n$ among rank $q$ projection matrices is attained when $\mathbf{P}_q$ is the orthogonal projector $\sum_{j=1}^q \mathbf{u}_{j, n} \mathbf{u}_{j, n}^\top$, in which case $\mathcal{R}_n\left(\mathbf{P}_q\right)=\sum_{j=q+1}^d \lambda_{j, n}$. In other words, batch PCA reduces to finding the first $q$ eigenvectors of the covariance $\boldsymbol{\Sigma}_n$ or, equivalently, the first $q$ singular vectors of $\mathbb{X}$. These eigenvectors are called principal components.

\subsubsection{Tuning parameters selection for the offline classifier}

\noindent According to the shape of our classifier presented in \eqref{p_est}, two tuning parameters, which are $K$ and $h_g$, have to be chosen by the user. It has been shown in the kernel smoothing estimation literature that the choice of the kernel $K$ does not significantly affect the quality of estimation of the nonparametric classifier $\widehat{\mathbb{P}}_{g,n}(\mathbf{x}_{\text{new}}).$ However, the choice of the bandwidth is a determinant factor of the quality of prediction. 

Here, we consider an optimal choice of the bandwidth for each new vector $\mathbf{x}_{\text{new}}$ according to the following formula:
\begin{eqnarray}\label{h}
h_{g,\text{opt}} = c_{g,\text{opt}} \max_{1\leq i \leq n} \|\mathbf{X}_i - \mathbf{x}_{\text{new}}\|\; n^{-\nu_{g,\text{opt}}},
\end{eqnarray}
where $c$ and $\nu$ are two positive constants such that $0 < c < 10$ and $0 < \nu < 1$ and obtained by minimizing the following cross-validation criterion calculated over the training sample:
\begin{eqnarray}\label{CV}
CV_g(c, \nu) := \dfrac{1}{n} \sum_{j=1}^n \left(\1_{\{Y_j = g \}} - \widehat{\mathbb{P}}_{g,n}^{(-j)}(\mathbf{X}_j) \right)^2,
\end{eqnarray}
where $\widehat{\mathbb{P}}_{g,n}^{(-j)}$ denotes the classifier's estimator based the training sample $(\mathbf{X}_1, Y_1), \dots, (\mathbf{X}_n, Y_n)$ leaving out the pair $(\mathbf{X}_j, Y_j).$ Then, $(c_{g,\text{opt}}, \nu_{g,\text{opt}})$ are such that
$$
(c_{g,\text{opt}}, \nu_{g,\text{opt}}) := \argmin_{(c, \nu)} CV_g(c, \nu).
$$

Note that the data-driven optimal bandwidth obtained in \eqref{h} is adaptive since its value changes with the new observation $\mathbf{x}_{\text{new}}.$ Moreover, once $h_{g,\text{opt}}$ is obtained, $Y_{\text{new}} = \arg\max_{g\in \mathcal{G}}  \widehat{\mathbb{P}}_{g,n}^{h_{g,\text{opt}}}(\mathbf{x}_{\text{new}})$, where $\widehat{\mathbb{P}}_{g,n}^{h_{g,\text{opt}}}(\mathbf{x}_{\text{new}})$ is the value of $\widehat{\mathbb{P}}_{g,n}(\mathbf{x}_{\text{new}})$ in \eqref{p_est} after replacing $h_g$ by $h_{g,\text{opt}}.$

\subsubsection{Algorithm for the offline classifier computation}

\noindent In this subsection we detail the algorithm that allows to predict the class of any out-of-sample observation. For generality purpose, we consider here that a Bach PCA is first applied on the training sample to reduce the dimension of the space of features. Therefore, the algorithm includes two step: (1) use Batch PCA to reduce the dimension of features and (2) calculate the offline classifier in \eqref{p_est} after choosing the optimal bandwidth according to \eqref{h} and replacing the original features ${\bf X}$ by the principle components obtained in step (1).

\begin{algorithm}[h]
\small{
\caption{Offline nonparametric supervised learning}
\label{CHalgorithm}
\begin{algorithmic}[1]
\STATE \textbf{Data:} a training sample of $n$ pairs $(\mathbf{X}_1, Y_1), \dots, (\mathbf{X}_{n}, Y_{n})$ copies of $(\mathbf{X}, Y)\in \mathbb{R}^d\times \mathcal{G}.$ 
\STATE\textbf{Batch PCA (dimension reduction step):} apply batch PCA of the set of data $\mathbb{X}=(\mathbf{X}_1, \dots, \mathbf{X}_n)$ to reduce the dimension of the space of predictors. The EVD allows to obtain the first $q$ orthonormal eigenvectors $\mathbf{u}_{1,n}, \dots, \mathbf{u}_{q,n}$. 
Then, the $q$ principle components are calculated such as $\mathbf{Z}_j := (Z_{1,j}, \dots, Z_{n,j})^\top = \mathbb{X}^\top \mathbf{u}_{j,n} \in \mathbb{R}^n$, for $j\in \{1, \dots, q\}.$ 
\STATE\textbf{Project} the new features $\mathbf{x}_{\text{new}}$ on the $q$-dimensional subspace, spanned by $\mathbf{u}_{1,n}, \dots, \mathbf{u}_{q,n}$, to finally obtain $\widetilde{\mathbf{x}}_{\text{new}} = (Z_{\text{new}, 1}, \dots, Z_{\text{new}, q})$ such that $Z_{\text{new},j} = \mathbf{x}_{\text{new}}^\top \mathbf{u}_{j,n}.$ 
\STATE \textbf{Offline classifier calculation step:}
\FOR{each $g$ in 1 {\bf to} G}
\STATE\textbf{Select optimal $(c, \nu)$:} use the new reduced training sample $(\widetilde{\mathbf{X}}_1, Y_1), \dots, (\widetilde{\mathbf{X}}_n, Y_n)$, where, for $i \in \{1, \dots, n\},$ $\widetilde{\mathbf{X}}_i = (Z_{i,1}, \dots, Z_{i,q})^\top \in \mathbb{R}^q$ and the cross-validation approach, detailed in \eqref{CV}, to find $(c_{g,\text{opt}}, \nu_{g,\text{opt}}).$ 
\STATE\textbf{Calculate optimal bandwidth for $\widetilde{\mathbf{x}}_{\text{new}}$}: use equation \eqref{h} to get $h_{g, \text{opt}}.$
\STATE {\bf Calculate $\widehat{\mathbb{P}}_{g,n}^{h_{g,\text{opt}}}$:} use equation \eqref{p_est}, where features $\mathbf{X}_i$ are replaced by $\widetilde{\mathbf{X}}_i$ and $h_g$ is replaced by $h_{g,\text{opt}}.$
\ENDFOR
\STATE\textbf{Class prediction}: $Y_{\text{new}} = \arg\max_{g\in \mathcal{G}}  \widehat{\mathbb{P}}_{g,n}^{h_{g,\text{opt}}}(\widetilde{\mathbf{x}}_{\text{new}}).$
\end{algorithmic}}
\end{algorithm}

\subsection{Online nonparametric supervised learning}\label{onlinesup}

\noindent In this subsection we discuss a fast algorithm that allows online estimation of the conditional probability $\widehat{\mathbb{P}}_{g,n}(\mathbf{x}_{\text{new}}).$ This algorithm is an adaptation of the Robbins-Monro algorithm (see \cite{RM}), based on stochastic approximation, to the supervised learning context. 

\subsubsection{Online nonparametric classifier estimation}
\noindent Given the definition in \eqref{proba}, one can say that $\mathbb{P}_g(\mathbf{x})$, for any fixed vector of features $\mathbf{x}$, is the regression function of the response variable $\1_{\{Y=g\}}$ on the vector of features $\mathbf{X}.$ Now, let $\left(\mathbf{X}_1, Y_1\right), \ldots,\left(\mathbf{X}_n, Y_n\right)$ be independent, identically distributed pairs of random variables, and let $f$ denote the joint probability density of $\mathbf{X}$. In order to construct a stochastic algorithm for the estimation of the classifier $\mathbb{P}_{g}: \mathbf{x} \mapsto \mathbb{E}(\1_{\{Y=g\}} | \mathbf{X}=\mathbf{x})$ at a fixed vector $\mathbf{x}$ such that $f(\mathbf{x}) \neq 0$, one may follow \cite{R} idea to define an algorithm, which approximates the zero of the function $M: a \mapsto f(\mathbf{x}) \mathbb{P}_{g}(\mathbf{x})-f(\mathbf{x}) a$. According to the Robbins-Monro procedure (see \cite{RM}), a recursive estimator of the nonparametric classifier, for a fixed vector of features $\mathbf{x}$ and for $n \geq 1,$ is defined as follows:
$$
\widehat{\mathbb{P}}_{g,n}(\mathbf{x})=\widehat{\mathbb{P}}_{g,n-1}(\mathbf{x})+\frac{1}{n} W_n(\mathbf{x}),
$$
where $W_n(\mathbf{x})$ is an ``observation" of the function $M(\cdot)$ at the point $\widehat{\mathbb{P}}_{g,n-1}(\mathbf{x})$. Considering $W_n(\mathbf{x}):= h_n^{-1} K(h_n^{-1}\|\mathbf{X}_n - \mathbf{x} \|)\left(\1_{\{Y_n=g\}} - \widehat{\mathbb{P}}_{g,n-1}(\mathbf{x})\right)$ one obtains the R\'ev\`ez-type estimator. 

It is worth noting that one may consider a general class of Robbins-Monro type estimators of the classifier $\mathbb{P}_g(\mathbf{x}_{\text{new}})$, for any fixed out-of-sample vector of features $\mathbf{x}_{\text{new}}$. That is
 \begin{eqnarray}\label{onlineclas}
    \widehat{\mathbb{P}}_{g,n}(\mathbf{x}_{new}) = \widehat{\mathbb{P}}_{g,n-1}(\mathbf{x}_{new})+\theta_n [\1_{\{Y_n=g\}}-\widehat{\mathbb{P}}_{g,n-1}(\mathbf{x}_{new})],
\end{eqnarray}
where $\theta_n >0$ is the step size that satisfies $\sum_{n=1}^\infty \theta_n = \infty$ and $\sum_{n=1}^\infty \theta_n^2 <\infty$ (\cite{RM} conditions). Note that in this article the initial value is calculated based on the offline estimator in equation \eqref{p_est} and using $n_0 << n$ observations from the available large-scale sample of size $n.$

\begin{rem}
The recursive, incremental, or online nonparametric classifier, presented in \eqref{onlineclas}, consists in updating the value of the classifier obtained at the step $n-1$ with a certain correction that depends on the freshly received observation at the step $n.$  The advantage of the recursive estimator of the classifier is that it does not necessarly require to store all the data at once on the same computer . Moreover, it is adapted to the data streaming context where data are received online. 
\end{rem}

\noindent{\it Discussion on the choice of the step size $\theta_n$:} In practice, several choices of $\theta_n$ are possible. Below, we provide a list of potential choices of $\theta_n$ then we discuss their advantages and disadvantages.
\begin{enumerate}
\item $\theta_n = 1/n.$
\item $\theta_n = \Delta_n /\sum_{i=1}^n \Delta_i (\mathbf{x})$, where $\Delta_n:= \Delta_n(\mathbf{x}):= K(\|\mathbf{X}_n-\mathbf{x} \|/h_n).$
\item $\theta_n =  \Delta_n h_n^d/\sum_{i=1}^n \Delta_i h_i^d,$ where $d$ is the dimension of $\mathbf{X}.$
\item $\theta_n = \Delta_n/nh_n^d$
\item $\theta_n = \dfrac{\gamma_n}{h_n^d} \Delta_n$, where $\gamma_n$ a well-chosen deterministic sequence.
\end{enumerate}

Note that a step size $\theta_n = 1/n$ is proportional to the sample size. Means that, more data are received less weight will be given to the correction term in \eqref{onlineclas}. The major drawback of this choice is that it does not depend on the out-of-sample vector of features $\mathbf{x}_{\text{new}}.$ Choices of step size depending on the local weighting function $\Delta_n$, the bandwidth $h_n$ and involving $\textbf{x}_{\text{new}}$ (as is the case for $\theta_n$ in (2)-(5)) seem more reasonable to consider. However, these step sizes request a pre-selection of an optimal bandwidth every time a new observation is received. Moreover, optimal bandwidths $(h_i)_{i=1, \dots, n}$ should be selected for all iterations up to time $n$ as is the case of $\theta_n$ in (2)-(3). If the selection of the bandwidth, at each iteration, is made through a cross-validation technique, the computation time required to calculate the online classifier will significantly increase. Note also that the choice of $\theta_n$ given in (4) leads to the R\'ev\`ez estimator which converges asymptotically when the bandwidth $h_n = n^{-\alpha}$, where $\alpha > 1/2.$ In such case the convergence rate is smaller that $n^{1/4}$ while the optimal convergence rate for kernel-type regression estimator, for $h_n = n^{-1/(d+4))}$, is $n^{2/(d+4)}.$ In the remaining of this paper we will consider a step size $\theta_n = \gamma_n h_n^{-1} \Delta_n$ which, consequently, leads to the generalized R\'ev\`ez estimator defined as follows:
 \begin{eqnarray}\label{revezgen}
    \widehat{\mathbb{P}}_{g,n}(\mathbf{x}_{new}) = \widehat{\mathbb{P}}_{g,n-1}(\mathbf{x}_{new})+ \theta_n [\1_{\{Y_n=g\}}-\widehat{\mathbb{P}}_{g,n-1}(\mathbf{x}_{new})],
\end{eqnarray}
where $\theta_n := \frac{\gamma_n}{h_n^d} \Delta_n.$

Moreover, the consistency of the generalized R\'ev\`ez estimator is guarantied for the following choices of bandwidth $h_n$ and $\gamma_n$. That is $\widehat{\mathbb{P}}_{g,n}(\mathbf{x}_{new})$ will converge, in probability, to $\mathbb{P}_{g}(\mathbf{x}_{\text new})$ with an optimal nonparametric rate of $n^{2/(d+4)}$ if
\begin{eqnarray}\label{convcond}
\gamma_n=\dfrac{c_\gamma}{n} \quad\text{and} \quad h_n=\dfrac{c_h}{n^{1/d+4}},
\end{eqnarray}
where $c_\gamma>0$ and $c_h>0.$ Note that the choice of $\gamma_n$ and $h_n$ in \eqref{convcond} allows to generalized R\'ev\`ez estimator to achieve an optimal nonparametric convergence rate of $n^{2/(d+4)}.$ Based on the choices made in \eqref{convcond}, and assuming, for simplicity, that $c_h=1,$ the step size becomes 
\begin{eqnarray}
    \label{theta_n5}
    \theta_n = \frac{c_{\gamma}}{n^{4/(d+4)}}\cdot K(n^{\frac{1}{d+4}}\cdot \|X_n-x\|).
\end{eqnarray}

\begin{rem}
It is worth noting here that $c_\gamma$ is optimally selected by the cross-validation technique. The quantity $c_\gamma$ does not depend on the sample size, and hence it is chosen only once.  
\end{rem}

Similarly to the offline nonparametric classifier estimation, the reduction of the dimension of the space of features is required to overcome the curse of dimension issue. This leads to discuss, in the next subsection, the online principle component analysis which can be used when the data are received in streaming or a massive database is available at hand.

\subsubsection{Online PCA}\label{opca} 

 In batch or offline PCA, EigenValues Decomposition (EVD) can be computed in $O(nd\min(n,d))$ floating points operations (flops), where $n$ is the sample size and $d$ the number of features (see \cite{GV}, Chap. 8). Regarding the memory allocation, batch PCA has at least $O(nd)$ space complexity as they require holding all the data in computer memory. Moreover, EVD requires $O(d^2)$ additional memory for storing the covariance matrix of dimension $d\times d.$ Therefore, the time and space complexity of offline PCA make it inefficient in the case massive data need to be analyzed or when the overall data structure changes when new observations are received in streaming. Nowadays, with the advanced progress in sensors and electronic devises, the size databases may increase exponentially as new records are received. Hence, it becomes very crucial to adopt a time-varying estimation of principle components through an online PCA for dimension reduction purpose.    

Several approached for online PCA were introduced in the literature to extend the bach PCA to the context of big data. Three main approaches were discussed in the literature. The first one is called perturbation methods (see \cite{GE}, \cite{G}, \cite{H}) that consist in perturbing the sample covariance $\boldsymbol{\Sigma}_n$ for each newly received observation. A recursive form of the vector of means and the sample covariance matrix are obtained as follows:
$
\boldsymbol{\mu}_{n+1}=\frac{n}{n+1} \boldsymbol{\mu}_n+\frac{1}{n+1} \mathbf{X}_{n+1},
$
and 
\begin{eqnarray}\label{sigma}
\boldsymbol{\Sigma}_{n+1}=\dfrac{n}{n+1}\boldsymbol{\Sigma}_n + \frac{n}{(n+1)^2}\left(\mathbf{X}_{n+1}-\boldsymbol{\mu}_n\right)\left(\mathbf{X}_{n+1}-\boldsymbol{\mu}_n\right)^T.
\end{eqnarray}
Then, if the eigenvalue decomposition of $\boldsymbol{\Sigma}_{n}$ is known, a natural idea to compute the eigenelements of $\boldsymbol{\Sigma}_{n+1}$ is to apply perturbation techniques to $\boldsymbol{\Sigma}_{n}$, provided that $n$ is sufficiently large and $n^{-1}\left(\mathbf{x}_{n+1}-\boldsymbol{\mu}_n\right)\left(\mathbf{x}_{n+1}-\boldsymbol{\mu}_n\right)^T$ is small compared to $\boldsymbol{\Sigma}_n.$ A major drawback of perturbation techniques for online PCA is that they require computing all eigenelements of the covariance matrix. Therefore, for each new received vector of features, $O(d^2)$ folps are needed to update the PCA and $O(d^2)$ memory space is required to store the results. 

The second approach is called reduced rank incremental PCA (IPCA) introduced in \cite{A} which is based on the incremental Singular Value Decomposition of \cite{B}. Compared to perturbation methods, IPCA does not require computing all $d$ eigenelements if the practitioner decides to keep only the largest $q < d$ eigenvalues into consideration. In the following, we summarize the IPCA approach. Consider $\boldsymbol{\Xi}_n = \mathbf{V}_n \mathbf{D}_n \mathbf{V}_n^\top$ a rank $q$ approximation to the covariance matrix $\boldsymbol{\Sigma}_n$, where $\mathbf{V}_n \in \mathcal{M}_{d,q}$ approximates the $q$ eigenvectors of $\boldsymbol{\Sigma}_n$ and $\mathbf{D}_n \in \mathcal{M}_{q,q}$ approximates its largest $q$ eigenvalues. When a new observation $\mathbf{X}_{n+1}$ is received, one updates $\boldsymbol{\Xi}_n$ as follows: (1) center the new vector $\overline{\mathbf{X}}_{n+1} := \mathbf{X}_n - \boldsymbol{\mu}_n$, then (2) decompose it as $\overline{\mathbf{X}}_{n+1} = \mathbf{V}_n \mathbf{c}_{n+1} + \overline{\mathbf{X}}_{n+1}^\perp$, where $\mathbf{c}_{n+1} = \mathbf{V}_n^\top \overline{\mathbf{X}}_{n+1}$ are the coordinates of $\overline{\mathbf{X}}_{n+1}$ in the $q$-dimensional space spanned by $\mathbf{V}_n$ and $\overline{\mathbf{X}}_{n+1}^\perp$ is the projection of $\overline{\mathbf{X}}_{n+1}$ onto the orthogonal space of $\mathbf{V}_n.$ Given equation \eqref{sigma}, the covariance matrix $\boldsymbol{\Sigma}_{n+1}$ is approximated by $\boldsymbol{\Xi}_{n+1} = \frac{n}{n+1}\boldsymbol{\Xi}_{n} + \frac{n}{(n+1)^2}\overline{\mathbf{X}}_{n+1} \overline{\mathbf{X}}_{n+1}^\top.$ Equivalently, one may write $\boldsymbol{\Xi}_{n+1}$ as follows:

$
\boldsymbol{\Xi}_{n+1}=\left[\mathbf{V}_n \frac{\overline{\mathbf{X}}_{n+1}^{\perp}}{\left\|\overline{\mathbf{X}}_{n+1}^{\perp}\right\|}\right] \mathbf{P}_{n+1}\left[\mathbf{V}_n \frac{\overline{\mathbf{X}}_{n+1}^{\perp}}{\left\|\overline{\mathbf{X}}_{n+1}^{\perp}\right\|}\right]^\top
$
and 
$
\mathbf{P}_{n+1}=\frac{n}{(n+1)^2}\left(\begin{array}{cc}
(n+1) \mathbf{D}_n+\mathbf{c}_{n+1} \mathbf{c}_{n+1}^\top & \left\|\overline{\mathbf{X}}_{n+1}^{\perp}\right\| \mathbf{c}_{n+1} \\
\left\|\overline{\mathbf{X}}_{n+1}^{\perp}\right\| \mathbf{c}_{n+1}^\top & \left\|\overline{\mathbf{X}}_{n+1}^{\perp}\right\|^2
\end{array}\right).
$

Finally, it suffices to perform the eigenvalue decomposition of the matrix $\mathbf{P}_{n+1} \in \mathcal{M}_{q+1, q+1}.$

The third approach is based on stochastic approximation algorithms and it encompasses stochastic gradient optimization method for online PCA (see \cite{S} for more details), and Candid covariance-free incremental PCA (CCIPCA) introduced in \cite{Wetal}. For further technical details about these methods the reader is referred to the review paper \cite{CD} where numerical comparison of the online PCA methods was conducted. The authors concluded that the IPCA and CCIPCA provide the most interesting trade-off between computation speed and statistical accuracy. 

\subsubsection{Computation of the online classifier} 

The following algorithm provides steps to classify any out-of-sample observation. The first step consists in using an online PCA to reduce the dimension of original features. Then, apply \eqref{revezgen} to calculate, recursively, the online classifier.

\begin{algorithm}[h]
\caption{Online nonparametric supervised learning}
\label{CHalgorithm}
\begin{algorithmic}[1]
\STATE\textbf{Data:} a training sample of $n$ pairs $(\mathbf{X}_1, Y_1), \dots, (\mathbf{X}_{n}, Y_{n})$ copies of $(\mathbf{X}, Y)\in \mathbb{R}^d\times \mathcal{G}.$ \\
\STATE\textbf{Online PCA step:}
\STATE\textbf{Initialization:} Use the first $n_0 << n$ pairs $\{(\mathbf{X}_1, Y_1), \dots, (\mathbf{X}_{n_0}, Y_{n_0}) \}$ to perform a Batch PCA on the initial data $\mathbb{X}_{n_0}=(\mathbf{X}_1, \dots, \mathbf{X}_{n_0})$   
\FOR{each $i$ in $(n_0+1)$ \text{to} $n$}
\STATE \textbf{update} the SVD obtained in the initialization step by using the online PCA (IPCA) as discussed in Subsection \ref{opca} and obtain the updated first $q$ orthonormal eigenvectors $\mathbf{u}_{1,i}, \dots, \mathbf{u}_{q,i}$. 
\ENDFOR
\STATE \textbf{Calculate} the $q << d$ first principle components such as $\mathbf{Z}_j := (Z_{1,j}, \dots, Z_{n,j})^\top = \mathbb{X}^\top_n \mathbf{u}_{j,n} \in \mathbb{R}^n$, for $j\in \{1, \dots, q\}.$ Then project the new features $\mathbf{x}_{\text{new}}$ on the final $q$-dimensional subspace, spanned by $\mathbf{u}_{1,n}, \dots, \mathbf{u}_{q,n}$, to finally obtain $\widetilde{\mathbf{x}}_{\text{new}} = (Z_{\text{new}, 1,n}, \dots, Z_{\text{new}, q, n})$ such that $Z_{\text{new},j,n} = \mathbf{x}_{\text{new}}^\top \mathbf{u}_{j,n}.$ 
\STATE \textbf{Online classifier calculation step:}
\FOR{each $g$ in 1 {\bf to} G}
\STATE\textbf{Initialization:} Use the first $n_0$ pairs $\{(\mathbf{X}_1, Y_1), \dots, (\mathbf{X}_{n_0}, Y_{n_0}) \}$ to calculate $\widehat{\mathbb{P}}_{g,n_0}(\widetilde{\mathbf{x}}_{\text{new}})$ using the offline estimator given in \eqref{p_est}.
\FOR{each $i$ in $(n_0+1)$ \text{to} $n$}
\STATE \textbf{Calculate $\widehat{\mathbb{P}}_{g,i}(\widetilde{\mathbf{x}}_{\text{new}})$:} use the recursive estimator in \eqref{revezgen} where the step size is as defined in \eqref{convcond}.
\ENDFOR
\ENDFOR
\STATE\textbf{Class prediction}: $Y_{\text{new}} = \arg\max_{g\in \mathcal{G}}  \widehat{\mathbb{P}}_{g,n}(\widetilde{\mathbf{x}}_{\text{new}}).$
\end{algorithmic}
\end{algorithm}

\section{Real-time monitoring of fetal well-being during pregnancy}\label{app}

\subsection{Data description}

Machine learning has made significant strides in obstetrics and improved the standard of care for expectant mothers and their babies. It is essential in the real-time monitoring of fetal health during pregnancy. Statistical learning algorithms can identify subtle patterns and deviations that may point to potential abnormalities or complications by analyzing extensive data collected from monitoring devices, such as fetal heart rate and maternal vital signs. 
This early detection enables medical professionals to take immediate action, reducing risks and improving outcomes. Additionally, machine learning models incorporate various factors, such as demographic data, medical history, and real-time monitoring data, to provide personalized risk assessments, allowing for tailored insights and well-informed decision-making. These algorithms are also excellent at identifying preterm births in advance, using large datasets to identify contributing factors and provide early warning signs. 

Moreover, machine learning is an effective tool for decision support, continuously analyzing monitoring data to offer healthcare providers insights and suggestions. In addition to improving prenatal care, this data-driven approach also stimulates obstetrics research and innovation, resulting in improved monitoring methods, fresh interventions, and a better understanding of fetal well-being. The application of machine learning algorithms for fetal well-being monitoring was recently considered in the literature. For instance, \cite{Ba} provided a review of several machine learning algorithms used to predict the fetal heart activity using signals produced by the fetal heart. \cite{Pradhan} used algorithms such as random forest, logistic regression, $k$NN and Gradient Boosting Machine to predict the fetal state. 

Cardiotocography (CTG) is a technique that aims to collect a sequence of measurements of the fetal heartbeat along with uterine contractions. The purpose of this technique is to monitor fetal well-being during pregnancy and labor. Before or during birth, babies may suffer from Oxygen deprivation for multiple reasons. This lack of Oxygen affects the growth of the baby’s organs and causes permanent damage to them. Hence, it is necessary for medical doctors to collect such information in order to detect the issue earlier and take timely actions. The available dataset here consists of 2126 fetal on which  21 features (variables) were recorded. The response variable here is the fetal state that takes value in one of the following categories $\mathcal{G}=$ \{Normal, Suspect, or Pathologic\}. Table \ref{ctg_variables} provides a description of the variables. 

\begin{table}[h]
\caption{CTG variables.}
\begin{tabular}{ll|ll} 
\hline
 Variables & Attribute Information & Variables & Attribute Information  \\
\hline
  LB & FHR baseline (beats per minute) & AC & number of accelerations per second \\
  FM & nbr.  of fetal movements per sec. & UC & nbr  of uterine contractions per sec. \\
  DL & nbr  of light decelerations per sec. & DS & nbr  of severe decelerations per sec. \\
DP & nbr  of prolonged decelerations per sec. & ASTV & \% of time with abnormal short term variability \\
MSTV & mean value of short term variability & ALTV & \% of time with abnormal long term variability \\
MLTV & mean value of long term variability  & Width & width of FHR histogram \\
Min & minimum of FHR histogram  & Max & maximum of FHR histogram  \\
Nmax & nbr.  of histogram peaks & Nzeros & nbr. of histogram zeros \\
Mode & histogram mode & mean & histogram mean \\
Median & histogram median & Variance & histogram variance \\
NSP & Fetal state: normal, suspect, Pathologic & & \\
\hline
\end{tabular}
\label{ctg_variables}
\end{table}

\subsection{Data exploration}
\begin{figure}[h]
  \begin{subfigure}{.5\textwidth}
    \includegraphics[width=1.25\linewidth]{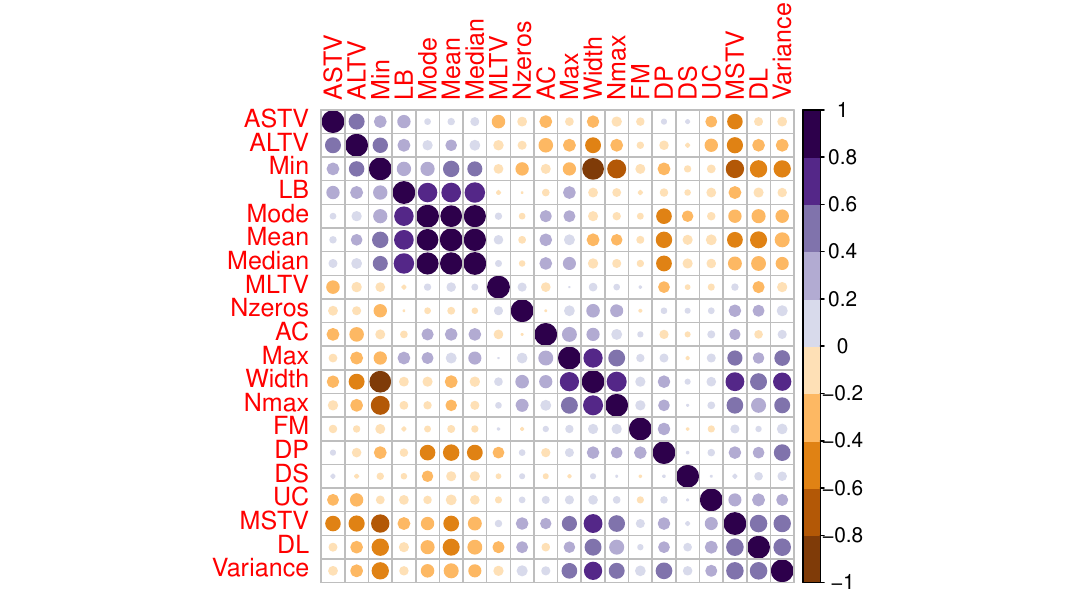}
  \end{subfigure}%
  \begin{subfigure}{.5\textwidth}
    \includegraphics[width=1.35\linewidth]{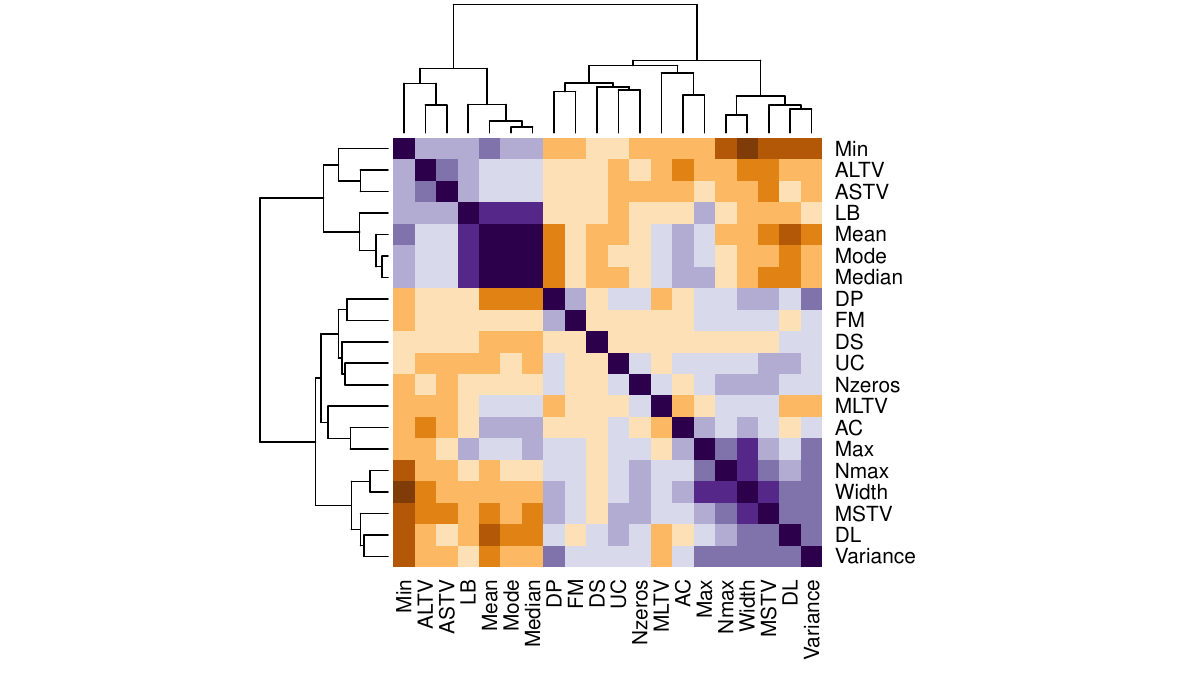}
  \end{subfigure}
  \caption{(Left) Correlation matrix of features in the CTG data. (Right) Heatmap of features in the CTG data.}
  \label{ctg_corrplot}
\end{figure}

Let us first have some insights about the CTG data. Figure \ref{ctg_corrplot} displays the correlation matrix of the features used to predict the fetal status. The heatmap of all features in the CTG data is presented in Figure \ref{ctg_corrplot}. Hierarchical clustering shows three groups of features. The first one is the group of positively correlated features including MIN, ALTV, ASTV, LB, Mean, and median. Strong positive correlation is noticed amoung the variables mean, mode, median and LB. The second group shows a weak negative correlation between DP, FM, DS, UC, NZeros, MLTV, AC, max and Nmax. Finally, strong negative correlation among the variables width, MSTV, DL, and variance.

As we have discussed above, dealing with high-dimensional data requires performing some dimension-reduction techniques to avoid the curse of dimension effect specifically when using kernel-based nonparametric classifiers. Figure \ref{ctg_corrplot} shows that the variables exhibit some strong correlations, which suggests using Batch PCA to extract the major important principle components that give a significant projection of the features in CTG data. 

\begin{figure}[h]
\centering
\includegraphics[width=4in, height=5in, clip,keepaspectratio]{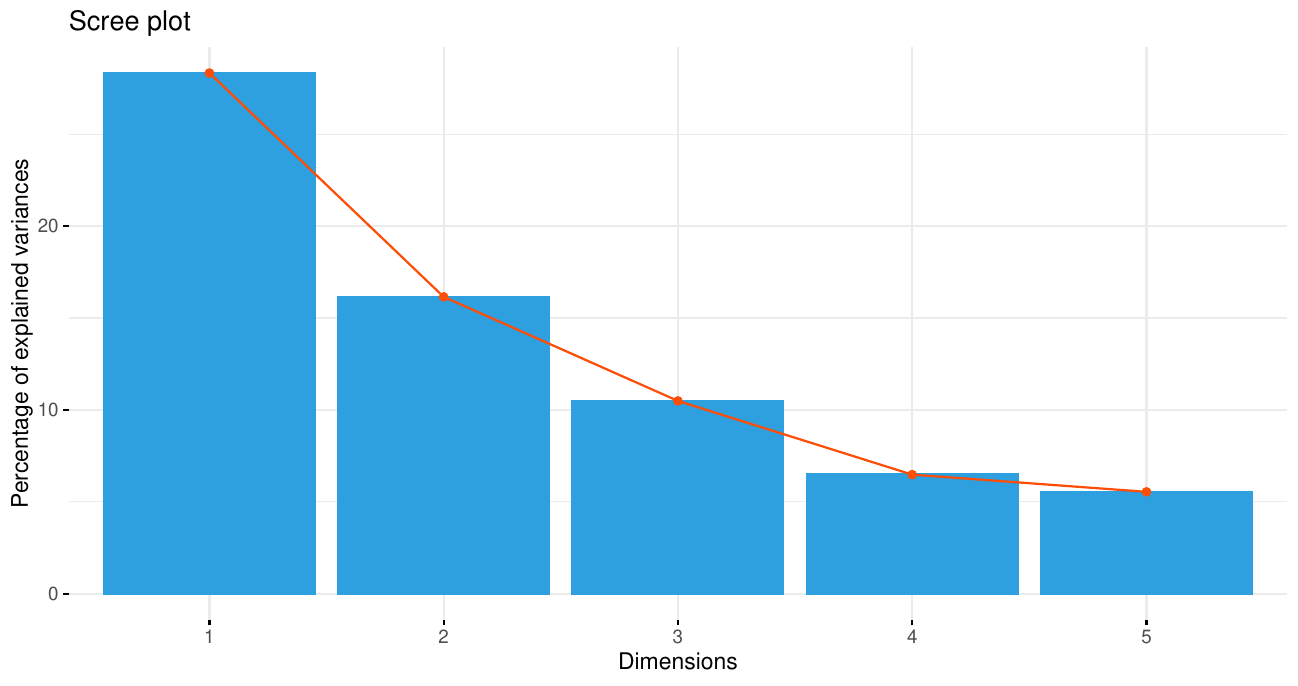} 
\caption{Percentage of explained variance obtained from Batch PCA.} 
\label{bar}
\end{figure}

The Factorial Analysis of Mixed Data (FAMD), which is a combination of PCA and Multiple Correspindence Analysis (MCA), confirms the correlation analysis described above. Figure \ref{bar} shows that the first two principle components of a batch PCA, made on quantitative features in CTG data, explain 44.4\% of the overlall variance in the data.

\begin{figure}[h]
\centering
\includegraphics[width=16cm, height=9cm]{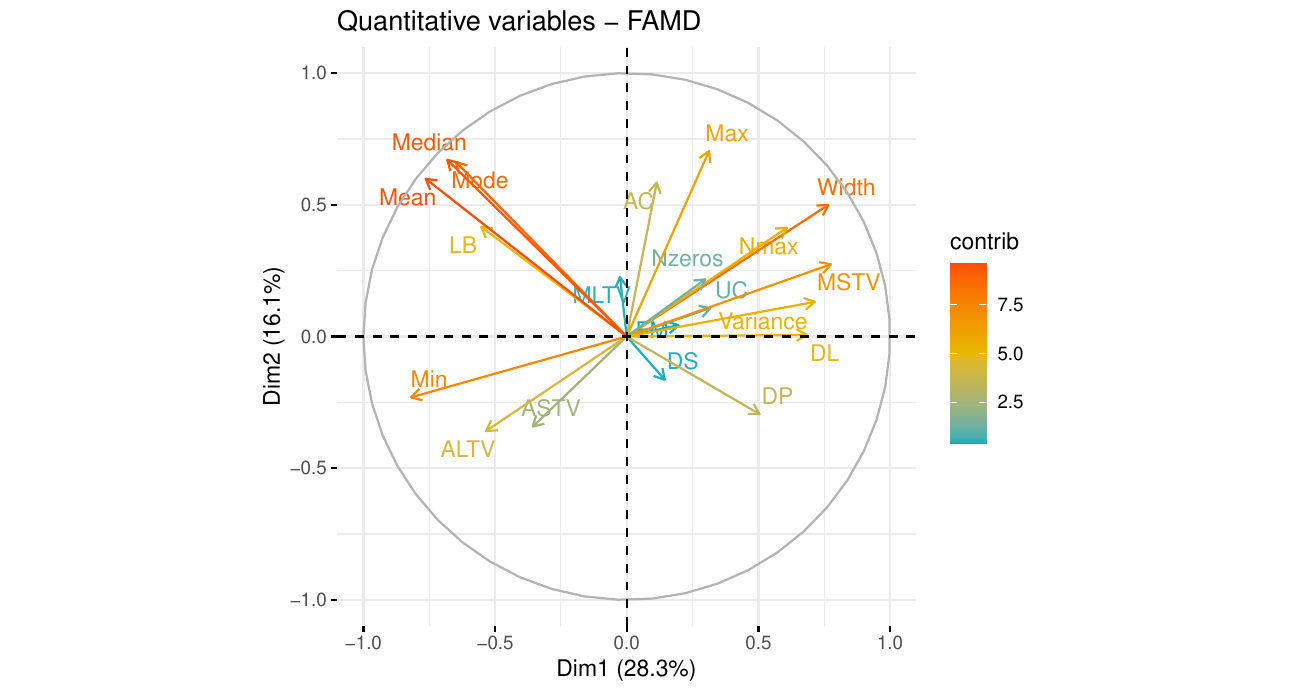}  
\caption{Projection of features on the first two principal components.} 
\label{variables}
\end{figure}

From Figure \ref{variables} one observes that the projection of features on the first two principle components shows three major group of variables well-projected in the factorial plan. The first group includes meadian, mean, mode and LB. The second one contains min, ALTV and ASTV. Finally the third group involves max, width, MSTV, Nmax, variance, DL, and AC. Note that the variables DS, MLTV, Nzeros, and FM are not well presented by the first two principle components.  

\begin{figure}[h]
\centering
\includegraphics[width=12cm, height=7cm]{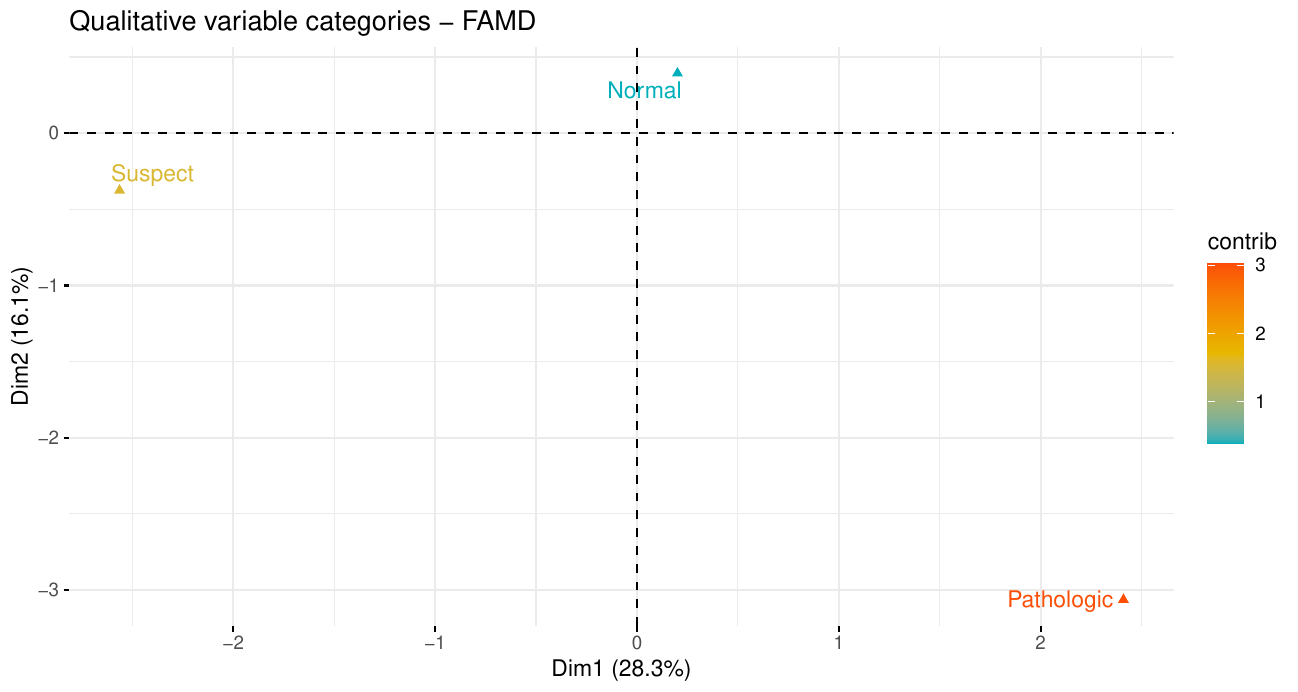}  
\caption{Screen Plot of the principal components of the CTG data.} 
\label{categorical}
\end{figure}

On the other side, Figure \ref{categorical} shows that the projection of the qualitative variable, which is fetal status, on the first two principle components shows that the first principle component opposes Suspect fetal to Normal ones. The second principle component contrasts the Normal fetal to the Pathologic ones. Note that the above offline factorial analysis is made based on all the available sample of 2126 observations. Figure \ref{bar} shows that, based on Batch PCA, over 66\% of the variability in CTG features is explained by the first five principle components. 

\subsection{Results}

In this subsection we want to compare the performance of the online classifier to the offline one using the CTG data. We use, as benchmark Four well-know supervised learning techniques commonly used in the literature, which are LDA, QDA, $k$NN and Random Forest (RF). 
\subsubsection{Choice of tuning parameters}
\begin{figure}[h]
\centering
\includegraphics[width=10cm, height=10cm, clip,keepaspectratio]{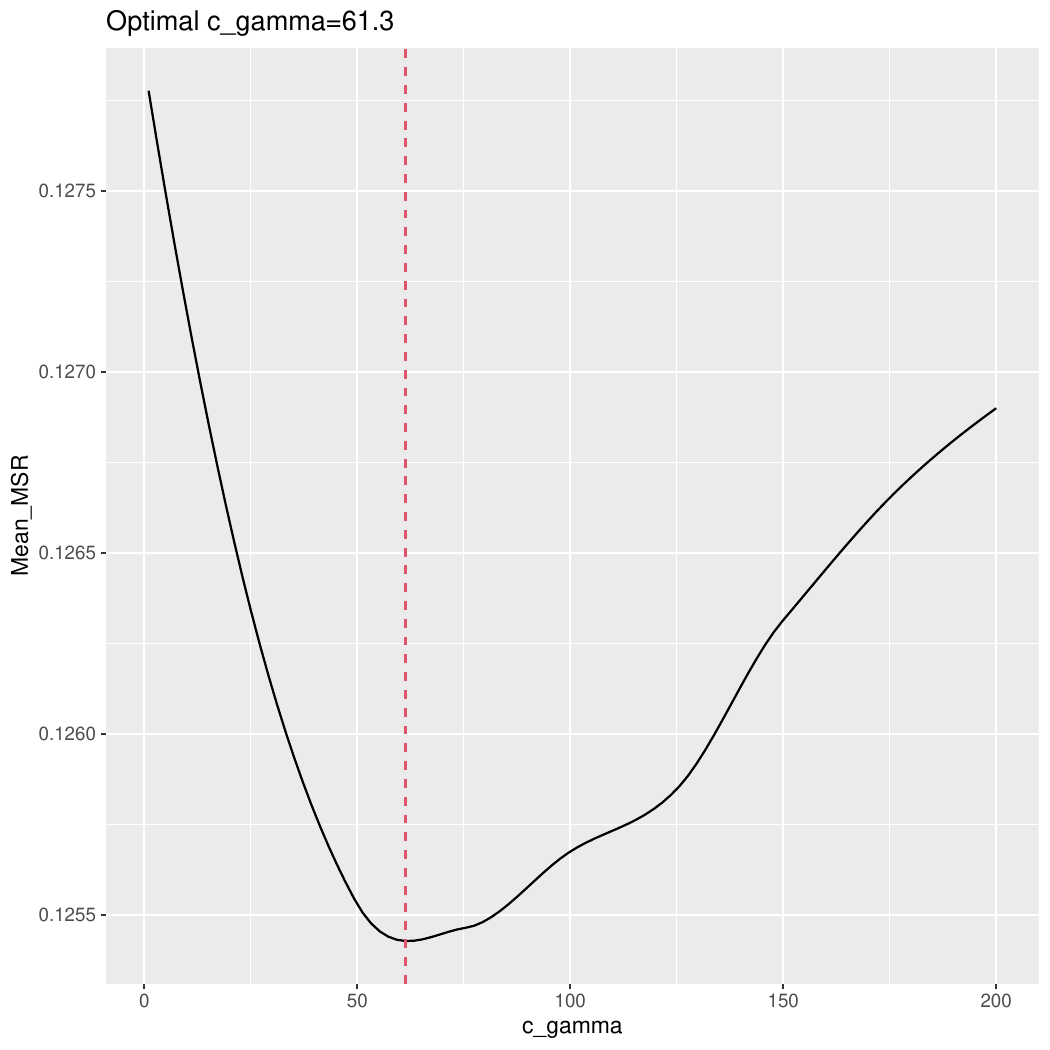} 
\caption{Optimal choice of $c_\gamma$ for online classifier.} 
\label{ctg_MSR_smooth}
\end{figure}

For the online classifier we consider the step size $$\theta_n = {c_{\gamma} n^{\frac{-4}{d+4}}\cdot K(n^{\frac{1}{d+4}}\cdot \|X_n-x\|)}.$$ Note that this form of step size does not require the computation of the optimal bandwidth and hence it is more efficient in reducing the computation time. On the other side the practitioner has to choose the hyperparameter $c_\gamma$. This parameter is estimated only once based on the first $n_0$ observations in the training sample $(\mathbf{X}_1,Y_1), \dots, (\mathbf{X}_{n_0}, Y_{n_0})$ used to calculate the initial values in the online PCA as well as the online classifier. The optimal $c_\gamma$ will be the one minimizing the misspecification rate calculated over the first $n_0$ observations. Figure \ref{ctg_MSR_smooth} shows that, for the CTG data, the optimal value of $c_\gamma$ is 61.3. The second tuning parameter is the kernel $K$ which is considered here to be the Epanishnikov kernel defined as $K(u) = \frac{3}{4}(1-u^2)\1_{(0,1)}(|u|).$ The choice of the kernel does not really affect the accuracy of the kenel-based classifiers. The Epanishnikov kernel will be considered to build the offline as well as the online classifiers.

Regarding the offline classifier, the choice of the bandwidth $h$ is crucial to obtain accurate predictions. In this paper, we adopt the cross-validation procedure described in \eqref{h} and \eqref{CV} to select the optimal bandwidth for the offline classifier. For the $k$NN classifier, the optimal number of neighbors $k$ is selected using cross-validation procedure where the loss function is misspecification rate.

\subsubsection{Performance assessment of classifiers}\label{perf}
To assess the performance of the above described classifiers, we split the data into training and testing subsamples according to the fetal state categories. The partition is given in Table \ref{NSP_training_test}.

\begin{table}[h]
\begin{center}
\caption{The partitioning of the sample by the NSP variable. }
\begin{tabular}{|c|c|c|c|c|} 
\hline 
\multicolumn{1}{|c|}{Fetal State} &
  \multicolumn{1}{c|}{Normal} & \multicolumn{1}{c|}{Suspect}  & \multicolumn{1}{c|}{Pathologic}  & \multicolumn{1}{c|}{Total}  \\
\hline 
Training sample & 1153 & 205 & 130 & 1488 \\
Test sample & 502 & 90 & 46 & 638\\
\hline
Total & 1655 (77.84\%) & 295 (13.87\%) & 176 (8.27\%) & 2126\\
\hline
\end{tabular}
\label{NSP_training_test}
\end{center}
\end{table}

\begin{figure}[h]
\centering
\includegraphics[width=10cm, height=10cm, clip,keepaspectratio]{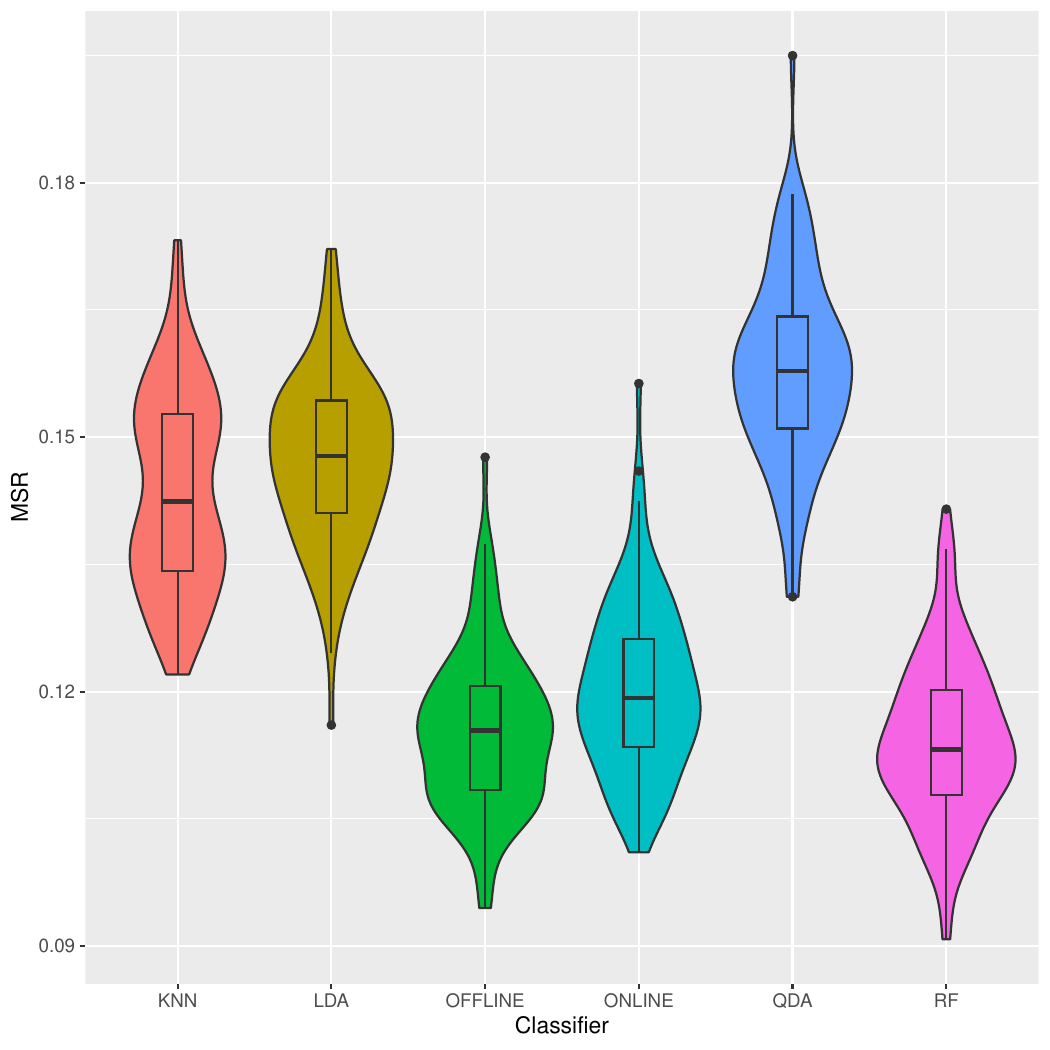} 
\caption{Violin plot and Boxplots of the misspecification rate (MSR) of each classifier.} 
\label{ctg_boxplot}
\end{figure}

Then, in order to assess the consistency (or robustness) of the classification results, we consider $M=500$ replications following the same learning/testing subsamples structure in Table \ref{NSP_training_test}. The first criterion used to compare between classifiers is the misspecification rate (MSR) defined as follows:
\begin{eqnarray}
    \label{MSR}
    \text{MSR}_k = \frac{1}{638}\sum_{i=1}^{638} \1_{\{Y_{i,k} \neq \widehat{Y}_{i,k} \}},\quad k=1, \dots, M,
\end{eqnarray}
where $Y_{i,k}$ is the class of the $i$th fetal in the testing sample obtained at the $k$th replication and $\widehat{Y}_{i,k}$ is its prediction using either LDA, QDA, $k$NN, RF, offline classifier or online classifier. Figure \ref{ctg_boxplot} displays the misspecification rates obtained over $M=500$ replications $(\text{MSR}_k)_{k=1, \dots, M}.$ Note that for the LDA, QDA, $k$NN and RF methods we first reduced the dimension of the space of features using online PCA then applied these techniques to predict the fetal status class. 

Figure \ref{ctg_boxplot} displays the viloin plot as well as the boxplot of misspecification rates and Table \ref{sumstat} shows the distribution of the $(\text{MSR}_k)_{k=1, \dots,M}$. One can notice that Random forest algorithm, the offline classifier and the online classifier consistently perform better than all other classifiers. Indeed, these three algorithms are giving very similar median misspecification rate of about $11.31\%, 11.54\%$ and $11.92\%$, respectively. Note that, based on the same dataset, \cite{Pradhan} found that, in terms of accuracy, Random Forest algorithm also performs better than the logistic regression, $k$NN and Gradient Boosting Machine.

\begin{table}[h]
\begin{center}
\caption{Distribution of the misspecification rates (in $\%$) obtained from 500 replications. }
\begin{tabular}{|l|c|c|c|c|c|c|c|} 
\hline 
& LDA & QDA & $k$NN &  Online & Offline & RF  \\
\hline
First quartile & 14.10 & 15.10 & 13.42  &  11.35 & 10.83 & 10.78  \\
Median & 14.77 & 15.78 & 14.24 &  11.92 & 11.54  & 11.31 \\
Mean & 14.72 & 15.78 & 14.36   & 12.04 & 11.53 & 11.43  \\
Third quartile & 15.43 & 16.42 & 15.27  & 12.62 & 12.06 & 12.02   \\
\hline
\end{tabular}
\label{sumstat}
\end{center}
\end{table}

In the following, we investigate the performance of each algorithm based one (not necessarily the optimal one) randomly chosen iteration out of the 500. We consider three comparison criterion commonly used in supervised learning to assess the overall as well as within each class of fetal state performance. First, the Recall (also named sensitivity) allows to know, when the actual class is $g$, how often the classifier was able to predict it correctly. Then, the Specificity which quantifies, when the actual class is not $g$, how often the classifier was able to predict is as not $g.$ The Balanced accuracy, for each class $g\in \mathcal{G}$, is the arithmetic mean between its Recall and its Specificity values. Note that, as shown in Table \ref{NSP_training_test}, the data are completely imbalanced. Indeed, Normal class represents $77.84\%$ of the data, Suspect class $13.87\%$ and Pathologic only $8.27\%$. In such case it is more appropriate to use Balanced Accuracy as a measure of performance rather than Recall or Specificity only.

Table \ref{performance} shows that the highest recall for Normal class is obtained with the offline classifier ($98\%$) followed by the RF algorithm (97.54\%). Online classifier gives the highest recall for the class Pathologic ($59.82\%$) and RF is the classifier with the highest recall for the class Suspect ($80.88\%$). In other words, when the fetal is Pathologic the online classifier was able to correctly detect it in $59.82\%$ of the cases and when it is Suspect the RF algorithm predict is correctly in $80.88\%$ of the cases. Note that for this case study (fetal well-being monitoring) we want the machine learning tool to have high performance in detecting Patholigic as well as Suspect cases. Indeed, an early detection of Pathologic or Suspect cases will allow the medical doctor to take the appropriate measures/treatment to save the fetal. For these two classes one can observe that RF and online classifier are giving the best performance. Moreover, this performance could significantly be improved if we had more data within these two classes for a better training. For the Specificity, the best performance is given by offline classifier for Suspect class ($99.47\%$), RF classifier for Pathologic class ($97.35\%$), then online classifier for Normal class ($73.89\%$). The Balanced Accuracy measure shows that RF algorithm is overall well performing for Normal class ($84.05\%$) and Suspect class ($89.85\%$), whereas online classifier is given a better result for Pathologic class ($77.68\%$). 

Table \ref{computationtime} displays the computation time (ins second) needed to train the algorithm and classify all fetals in the testing subsample. It also reports the accuracy and the F1-score as measure of overall performance. One can observe that the online and offline classifiers as well as RF algorithm give the best performance vis-a-vis Accuracy and F1-Score. All computation made in this paper was based on a personal computer (Apple MacBook Air, M2 chip, with 8GB of RAM). In terms of computation cost, RF is the fastest, followed by the online classifier. It is worth noting here that the Random Forest algorithm was performed using \texttt{ranger} packages in R software which is designed to build Random Forest models efficiently, particularly for large datasets. It leverages parallel computation and optimized algorithms to train models quickly while maintaining high accuracy. This gives advantage of the RF algorithm with respect to the other methods. Finally, one can observe that the time required by the offline classifier is almost 15 times the one needed by the online classifier.

\begin{table}[h]
\begin{center}
\caption{Some measures of performance for each classifier. }
\begin{tabular}{|l|l|c|c|c|} 
\hline 
\multicolumn{1}{|l|}{} & \multicolumn{1}{|c|}{Class} &
  \multicolumn{1}{c|}{Recall (\%)} & \multicolumn{1}{c|}{Specificity (\%)} & \multicolumn{1}{c|}{Balanced Accuracy (\%)}    \\
\hline 
& Normal & 95.70 & 65.00 & 80.35   \\
LDA & Pathologic& 8.9   & 91.65 & 46.27 \\
& Suspect& 16.17 & 90.43 & 53.30 \\
\hline
& Normal & 93.09 & 69.44 & 81.27  \\
QDA & Pathologic& 57.14 & 94.85 & 75.99\\
& Suspect& 67.64 & 96.98 & 82.31\\
\hline
& Normal & 95.70 & 71.11 & 83.41 \\
KNN & Pathologic& 58.92 & 94.43 & 76,67 \\
& Suspect& 69.11 & 98.55 & 83.83\\
\hline
& Normal & 97.54 & 70.56 & {\bf 84.05} \\
RF & Pathologic& 53.57 & {\bf 97.35} & 75.46  \\
& Suspect& {\bf 80.88} & 98.82 & {\bf 89.85}\\
\hline
& Normal & 94.01 & {\bf 73.89} & 83.95  \\
Online & Pathologic& {\bf 59.82} & 95.54 & {\bf 77.68} \\
& Suspect& 72.05 & 97.90 & 84.98 \\
\hline
& Normal & {\bf 98.16} & 65.56 & 81.86  \\
Offline & Pathologic& 49.10 & 96.94 & 73.02 \\
& Suspect& 72.05 & {\bf 99.47} & 85.76 \\
\hline
\end{tabular}
\label{performance}
\end{center}
\end{table}




\begin{table}[h]
\begin{center}
\caption{CPU time, accuracy and F1-score obtained on the testing subsample. }
\begin{tabular}{|l|c|c|c|c|c|c|} 
\hline 
& LDA & QDA  & $k$NN & Online  & Offline & RF \\
\hline 
CPU time (second) & 0.046 & 0.027 & 16.84  &  31.75 & 471.93 & 0.584\\
Accuracy (\%) & 76.41 & 85.6 & 87.61 &  88.57 & 89.41 & 90.25\\
F1-score & 0.931 & 0.923 & 0.934 &  0.939 & 0.945 & 0.948\\
\hline
\end{tabular}
\label{computationtime}
\end{center}
\end{table}

\subsubsection{More focus on the comparison between offline and online classifiers}

The purpose here is to focus more on the comparison of the performance offline and online classifier using other criterion than those used in Subsection \ref{computationtime}. 

\begin{figure}[h]
	\centering
	\begin{subfigure}{0.32\linewidth}
		\includegraphics[width=\linewidth]{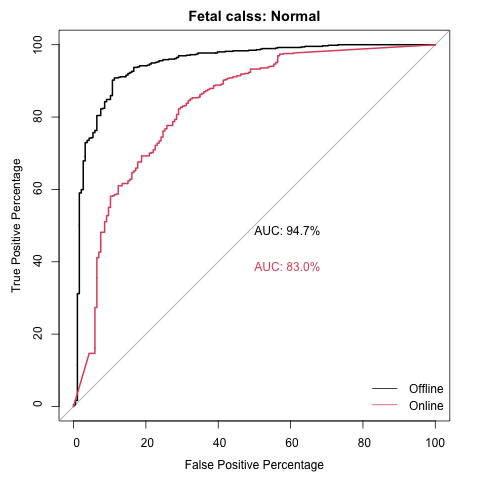}
		\caption{}
		\label{fig:subfigA}
	\end{subfigure}
	\begin{subfigure}{0.32\linewidth}
		\includegraphics[width=\linewidth]{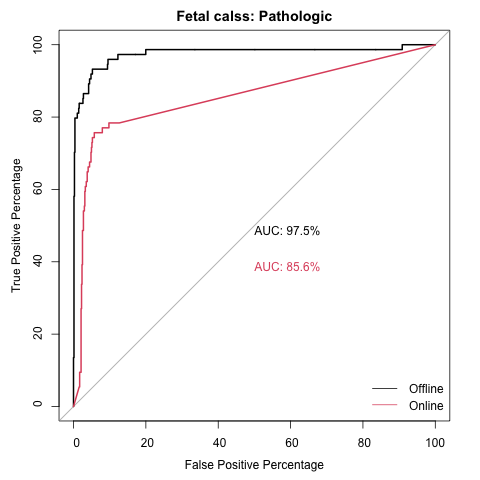}
		\caption{}
		\label{fig:subfigB}
	\end{subfigure}
	\begin{subfigure}{0.32\linewidth}
	        \includegraphics[width=\linewidth]{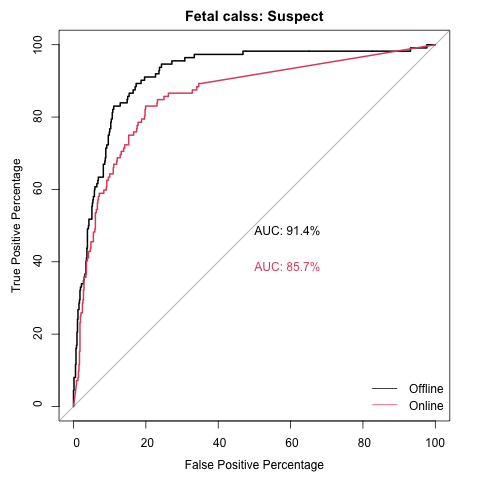}
	        \caption{}
	        \label{fig:subfigC}
         \end{subfigure}
	\caption{ROC and AUC for offline and online classifiers for fetal Status: Normal (a), Pathologic (b) and Suspect (c).}
	\label{ROC}
\end{figure}

Figures \ref{ROC} displays the receiver operating characteristic curve (ROC) curve as well as the Area under the ROC Curve (AUC) for the offline and online classifiers obtained for the fetal status class Normal, Suspect and Pathologic. One can notice that the offline classifier performs better than the online one. Indeed, for Normal class the offline classifier has an AUC equals to $94.7\%$ against $83\%$ for the online classifier. Similarly, for the Pathlogic (resp. Suspect) class, the offline classifier given an AUC of $97.5\%$ (resp. $91.4\%$), while the online classifier has an AUC of $85.6\%$ (resp. $85.7\%$). The offline classifier requires a large computation time (almost 15 times the time used by the online classifier) because of the selection of the optimal which is also made using cross-validation method. Finally, compared to other methods, the online classifier gives the best computation-time/accuracy trade-off.

\section{Conclusion}

In this paper novel nonparametric supervised learning techniques are introduced to classify massive data. First, we introduced an offline kernel-based nonparametric classifier. This classifier has the advantage of not imposing any probability distribution restriction on the vector of feature. It is flexible (in terms of linearity) and can be easily learned directly from the data. However, the offline classifier suffers from the ``curse of dimension" issue. In other words this classifier's performance decreases drastically as soon as the dimension of features is large. For this a Bach PCA is used to reduce the dimension of the features and enhance the offline classifier's performance. Then, we introduced a nonparametric supervised classification algorithm adapted to the context of massive data or when observations are received in streaming. The online classifier is calculated in two steps. First we use an online PCA to reduce the dimension of features in real-time. Then, we use the Robbins-Monro approach to define a kernel-based recursive classifier. Application to real-time fetal well-being monitoring during pregnancy showed that the offline and online nonparametric classifiers have a performance compared to well-known machine learning algorithms such as Random Forest. Finally, we show that the online classifier gives the best computation-time/accuracy trade-off compared to the offline classifier.

\end{document}